%% file: NLP.tex
\documentclass[10pt,twocolumn,letterpaper]{article}

\usepackage{cvpr}              

\input{includes/krishnas_includes}

\input{includes/krishnas_macros}

\definecolor{Gray}{gray}{0.95}
\definecolor{highlight}{rgb}{0.824,0.976,0.824}
\definecolor{darkgray}{rgb}{0.902,0.898,0.902} 
\definecolor{dim}{rgb}{0.95,0.95,0.95}
\definecolor{uniform}{rgb}{0.702,0.702,0.702}
\definecolor{delta-color}{rgb}{0.824,0.824,0.976}
\definecolor{baseline}{gray}{0.9}

\definecolor{cvprblue}{rgb}{0.21,0.49,0.74}
\usepackage[pagebackref,breaklinks,colorlinks,citecolor=cvprblue]{hyperref}

\def\paperID{23} 
\def\confName{CVPR}
\def\confYear{2024}

\title{Response Wide Shut: Surprising Observations in Basic Vision Language Model Capabilities}
\author{
Shivam Chandhok$^{1,2}$ \quad Wan-Cyuan Fan$^{1,2}$ \quad Leonid Sigal$^{1,2,3}$ \vspace{2mm} \\
$^1$University of British Columbia \quad $^2$Vector Institute for AI \quad
$^3$CIFAR AI Chair \vspace{1mm} \\
\texttt{\small \{chshivam, wancyuan, lsigal\}@cs.ubc.ca}
}

\begin{document}


\maketitle
\thispagestyle{empty}

\begin{abstract}
Vision-Language Models (VLMs) have emerged as general purpose tools for addressing a variety of complex computer vision problems. 
Such models have been shown to be highly capable,
but, at the same time, also lacking some basic visual understanding skills.
In this paper, we set out to understand the limitations of SoTA VLMs on fundamental visual tasks: object classification, understanding spatial arrangement, and ability to delineate individual object instances (through counting), by constructing a series of tests that probe which components of design, specifically, maybe lacking. 
Importantly, we go significantly beyond the current benchmarks, that simply measure final performance of VLM, by also comparing and contrasting it to performance of probes trained directly on features obtained from visual encoder (image embeddings), as well as intermediate vision-language projection used to bridge image-encoder and LLM-decoder ouput in many SoTA models ({\em e.g.}, LLaVA, BLIP, InstructBLIP). In doing so, we uncover nascent shortcomings in VLMs response and make a number of important observations which could help train and develop more effective VLM models in future.
\end{abstract}
\vspace{-15pt}
\section{Introduction}
\label{sec:1}
Over the past couple of years Large Language Models (LLMs) and Multi-modal LLMs (MLLMs) have markedly emerged as general purpose models capable of addressing many language and computer vision understanding problems\cite{2023videochat,llava}. 
 While benchmarks like SEED-Bench-2 \cite{li2023seed} illustrate ability of LLMs and MLLMs to succeed at many complex tasks, {\em e.g.}, ability to comprehend charts, humor and memes in images and videos, at the same time there is also growing body of evidence that such models lack the very basic capabilities that are seemingly necessary to solve high-level tasks they are succeeding on ({\em e.g.}, fundamental lack of ability of GPT-like models to perform basic mathematical functions \cite{dziri2023faith}). Such observations seem to suggest that the mechanism for solving more complex tasks in LLMs may be markedly different from those in humans, relying more on large-scale matching and memory recall, as opposed to more functional and step-by-step reasoning. 
\begin{figure}[t]
    \centering
    \includegraphics[width=0.9\linewidth]{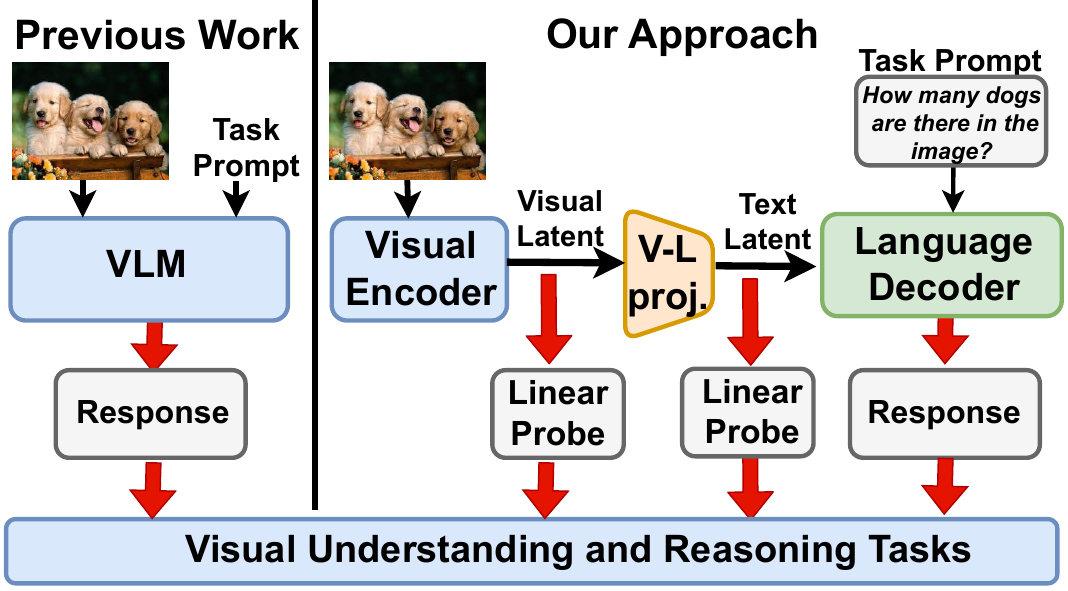} 
    \caption{ Different from previous work which analyse VLMs as a whole on given task (left), we propose to look at performance of VLMs in terms of intermediate spaces that represent knowledge as it is processed through the VLM network.\vspace{-0.3in}}
    \label{fig:intro_figure}
    \vspace{-0.1in}
\end{figure}

These seemingly competing and sometimes contradictory observations, in our opinion, require more detailed and systematic analysis and understanding.
Inspired from this, we propose to look at performance of MLLMs \textit{in terms of intermediate spaces that represent information as it is processed through the MLLM network} (see Figure \ref{fig:intro_figure}) Specifically, we make an observation that most MLLM models consist of image encoder, text decoder, and (optional) image-to-text projection/alignment mechanism.
Therefore, by analyzing these different spaces on their ability to perform fundamental visual tasks, we can better understand which capabilities may be missing and where. In this paper, we analyse a series of core vision capabilities that we posit would be required for any form of higher-level visual reasoning tasks. Specifically: (i) ability to recognize objects (coarse and fine-grained classification), (ii) understand their spatial arrangement, and (iii) delineate instances of a given object type (counting). 
Fundamentally, we ask the following question, how can the existing models be best used for various fundamental tasks and what should be the focus of future improvements in MLLM developments.

Through our detailed experimental design and analysis we make a number of important observations. Mainly, that (1) MLLMs are much less capable of recognizing fine-grained categories over the course-grained counterparts; while this, in itself, may not be surprising we illustrate that such drop in performance is largely attributable to vision-language projection \& alignment with the language decoder. The visual and text latents contain knowledge required to discriminate between fine-grained classes but this does not reach the final response space of VLM which shows drastically low performance.(2) Similarly, we also find that the visual and text latents have knowledge required for object counting but the response space performance is  relatively poor for counting task (3) Finally, we observe significant lack of ability of vision encoders to capture spatial arrangement information (similar to \cite{tong2024eyes}), which results in significant drop of performance for spatial understanding tasks. Interestingly, we find LLM decoders within VLMs that are instruction fine-tuned, bridge this performance gap somewhat by relying on embedded language priors. 
\begin{figure}[t]
    \centering
    \includegraphics[width=0.9\linewidth]{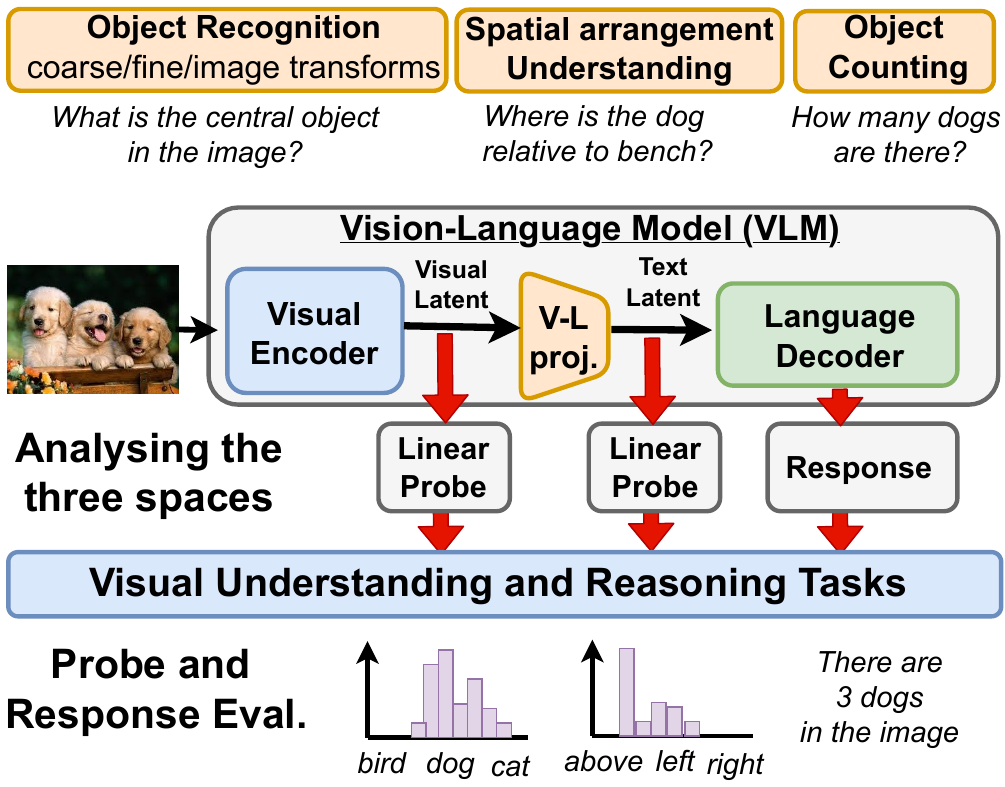} 
    \caption{{\bf Overview of our proposed approach for analysing visual understanding capabilities of VLMs.} Specifically, we analyse the three spaces within a VLM i.e visual latent, text latent and response space to get a nuanced understanding of what aspects of visual information are captured within VLMs, how visual knowledge flows through the network and where should improvements be targetted to alleviate the deficiencies of VLMs.\vspace{-0.2in}}
    \label{fig:main_figure}
    \vspace{-0.1in}
\end{figure}
\vspace{-10pt}
\section{Related Work} \label{sec:2}
Advances in multimodal vision and NLP have led to the development of diverse VLMs \cite{instructblip, li2023blip2, llava} and increased interest in understanding these models \cite{chen2023minigptv2, probing3d, distribution, finer, li2023seed, visual_datatypes, zhu2023minigpt}. 

{\bf Vision Language Models and their Categorization.} Broadly, vision-language foundational models can be categorized into three groups based on the specific objectives they optimize (training paradigm) and architecture design. (1) \textit{Contrastive Multi-Encoder models}: The models in this category employ a separate encoder for each modality ({\em i.e.}, image and text) and use some form of a weak contrastive alignment loss between image and text inputs for training. For e.g, CLIP \cite{clip}
(2) \textit{Encoder-Decoder Generative models}: The models in this category employ an encoder-decoder architecture to first project images to a low-dimensional representation and then use the representation as context to the language decoder to generate language descriptions or captions with generative language modeling loss. For eg, the BLIP-2 \cite{li2023blip2}.
(3) \textit{Instruction Fine-Tuned models}: These models are additionally fine-tuned to understand human instructions and intent while answering visual questions. Usually, these models are fine-tuned with some form or variant of RLHF \cite{lambert2022illustrating} or direct preference optimization which allows them to understand the kind of questions humans ask about visual content and the type of answers they expect. For e.g, LLaVA \cite{llava}, InstructBLIP \cite{instructblip}.

{\bf Analyzing Vision Language Models.} Recent interests in pre-trained VLMs have motivated research that aims to understand their visual capabilities. Han et al. \cite{distribution} investigate the object recognition and visual understanding capabilities of VLMs under diverse distribution shifts including (but not limited to) style changes, domain changes, and Gaussian noise. Specifically, they investigate the visual understanding capabilities of CLIP \cite{clip}, LLaVA \cite{llava}, and GPT4-Vision \cite{openai2023gpt4} models across varying types and levels of domain shift. Furthermore, Udandarao et al. \cite{visual_datatypes} investigate vision-language models by adding perturbation to images and analysing if VLMs inherently recognize the type of perturbation added to the image.
However, these works often focus on high-level analysis of VLM skills or lack there off; we focus on more detailed understanding by also looking at intermediate feature representations and diagnostic foundational visual understanding tasks.
\begin{table*}[h]
\caption{{\bf Course-grained Recognition Performance.} All models appear to be highly capable and capturing the information required at all stages of VLMs to make predictions. BLIP family of models do appear to suffer from either no, or lacking, instruction tuning which results in lower response.}
\small
\centering
\begin{tabular}{l c ccc ccc ccc}
\toprule
\multirow{2}{*}{VLM Method} & & \multicolumn{3}{c}{PaintSkills \cite{DallEval}} & \multicolumn{3}{c}{Pascal VOC \cite{Everingham10}} & \multicolumn{3}{c}{\bf Average} \\ 
\cmidrule(lr){3-5} \cmidrule(lr){6-8} \cmidrule(lr){9-11} 
& & Visual & Text & Resp.
& Visual & Text & Resp. 
& Visual & Text & Resp. \\  \midrule
CLIP \cite{clip} & & 99.2 & - & 98.0 & 97 & - & 95.0 & 98.1 & - & 96.5 \\ \hline
BLIP \cite{li2023blip2} & 7B & 99.2 & 99.2 & 60.0 & 97.8 & 98.8 & 38.0 & 98.5 &  99.0 &{49.0}\\
InstructBLIP \cite{instructblip} & 7B & 99.9 & 99.2 & 98.0 & 98.2 & 99.0 & 69.0 & 99.1 & 99.1 &{83.5} \\ \hline
\multirow{2}{*}{LLaVA \cite{llava}} & 7B & 99.2 & 99.2 & 97.0 & 96.3 & 95.5 & 91.0 & 97.8 & 97.4 & 94 \\ 
        & 13B & 99.8 & 99.8 & 97.7 & 96.4 & 95.0 & 93.0 & 98.1 & 97.4 &95.3 \\                            
\bottomrule
\end{tabular}
\end{table*}

\begin{table*}[h]
\caption{{\bf Fine-grained Recognition Performance.} 
There is significant drop in performance in the response space for fine-grain vs. course-grained recognition. In the intermediate visual and text feature spaces the drop also exists, but is significantly less comparatively.
\vspace{-15pt}}
\small
\centering
\begin{tabular}{l c ccc ccc ccc}
\toprule
\multirow{2}{*}{VLM Method} & & \multicolumn{3}{c}{Stanford Dogs \cite{dogs}} & \multicolumn{3}{c}{CUB \cite{CUB}} & \multicolumn{3}{c}{\bf Average} \\ 
\cmidrule(lr){3-5} \cmidrule(lr){6-8} \cmidrule(lr){9-11} 
& & Visual & Text & Resp.
& Visual & Text & Resp. 
& Visual & Text & Resp. \\  \midrule 
CLIP \cite{clip} & & 92.0 & - & 84.0 & 93.0 & - & 71 & 92.5 & - & 77.5\\ \hline
BLIP \cite{li2023blip2} & 7B & 93.6 & 92.5 & 23.5 & 92.0 & 93.2 & 9.5 & 92.5 &92.8 &16.5 \\
InstructBLIP \cite{instructblip} & 7B & 93.6 & 93.2 & 12.7 & 91.7 & 95.0 & 13.0 &92.6 & 94.1 & 12.9 \\ \hline
\multirow{2}{*}{LLaVA \cite{llava}} & 7B & 91.9 & 88.7 & 21.1 & 92.2 & 87.3 & 34 & 92 & 88 & 27.5 \\ 
        & 13B & 91.4 & 89.2 & 33.4 & 92.2 & 86.5 & 35.5 & 92 & 87.9 & 34.5 \\                            
\bottomrule
\end{tabular}
\end{table*}

\vspace{-10pt}

\section{Methodology}
\label{sec:method}
\subsection{Overview}
Motivated by the modular design of recent VLM models, where individual components are often pre-trained separately and later put together for end-to-end fine-tuning, we propose to investigate VLMs by looking at their performance in terms of intermediate spaces that represent information as it is processed through the VLM (Figure ~\ref{fig:main_figure})

Specifically, vision-language foundational models (VLMs), {\em e.g.}, BLIP \cite{li2023blip2}, InstructBLIP \cite{instructblip}, LLaVA \cite{llava}, typically comprise of a \emph{visual encoder}, \emph{vision-language projection} module and \emph{language decoder} (see Figure~\ref{fig:main_figure}). This gives rise to 3 different spaces within a VLM :-  (1) \emph{visual latent} (output of visual encoder) space; (2) \emph{vision-language shared latent} (output of vision-language projection) space, and (3) \emph{language response} space (output of language decoder). We conjuncture that these spaces might capture different (diverse) aspects of an visual content which cumulatively help a VLM understand visual content. While not all VLMs contain all three components ({\em e.g.}, CLIP only has visual latents and a variant of discrete decoder that leverages matching in the latent space), this provides a useful abstraction for the analysis. To this end, we probe these different spaces and separately evaluate/analyse their visual capabilities to get a nuanced understanding of what aspects of visual information are captured within VLMs and where.

\vspace{-15pt}
\paragraph{VLM Models and Experimental design:}
We analyse the visual capabilities of prominent open-source VLMs belonging to different families (as described in Section \ref{sec:2}) \textit{CLIP \cite{clip}, BLIP2 \cite{li2023blip2}, InstructBLIP \cite{instructblip} and LLaVA-1.5 \cite{llava}}. For intermediate spaces i.e visual and text latents (see Fig \ref{fig:main_figure}), we probe the frozen feature representations by training task specific linear probes and employing zero-shot inference procedures following previous work \cite{probing3d}. For response space, VLM is prompted with a VQA query such as ``{\em What is the central object in the image? Choices - dog, cat, ...?}" and it has to respond with the correct answer, following protocol introduced in previous work \cite{distribution}.  \\

\vspace{-25pt}
\paragraph{Choice of Diagnostic Datasets:}
PaintSkills \cite{DallEval} is a compositional diagnostic evaluation dataset recently developed to measure fundamental visual reasoning and understanding capabilities (i.e classification, counting, spatial arrangement) in foundational models. PASCAL VOC \cite{Everingham10} is  a standard dataset widely used in the computer vision community; it consists of real-world images for a set of coarse-grained classes. CUB \cite{CUB} and Stanford Dogs \cite{dogs} are widely used fine-grained datasets used for fine-grained recognition. For fair comparisons and analysis across datasets we consider a subset of $15$ classes per dataset to ensures that chance performance (and thus, task difficulty) across them is the same.

\begin{table}[h]
    \caption{{\bf Object Counting (top) and Spatial Understanding (bottom) performance on PaintSkills \cite{DallEval} dataset.} See text for detailed analysis and discussion. \vspace{-20pt}}
    \begin{subtable}[h]{0.5\textwidth}
        \small
        \centering
        \begin{tabular}{@{}lcccc@{}}
        \toprule
        \multirow{2}{*}{VLM Method} & & \multicolumn{3}{c}{PaintSkills \cite{DallEval}}  \\ 
        \cmidrule(lr){3-5}
        & & Visual & Text & Resp. \\  \midrule 
        CLIP \cite{clip} & & 93.5 & - & 49.0 \\ \hline
        BLIP \cite{li2023blip2} & 7B &  96.6 & 95.3 & 25.0 \\
        InstructBLIP \cite{instructblip}    & 7B &  96.6 & 95.6 & 82.0 \\ \hline
        \multirow{2}{*}{LLaVA \cite{llava}} & 7B &  93.0 & 94.0  & 80.0 \\ 
                                            & 13B & 93.1 &  94.0 & 79.8 \\                            
        \bottomrule
        \end{tabular}
    \end{subtable}
    \hfill
    \begin{subtable}[h]{0.5\textwidth}
        \small
        \centering
        \begin{tabular}{@{}lcccc@{}}
        \toprule
        \multirow{2}{*}{VLM Method} & & \multicolumn{3}{c}{PaintSkills \cite{DallEval}}  \\ 
        \cmidrule(lr){3-5}
        & & Visual & Text & Resp. \\  \midrule 
        CLIP \cite{clip} & & 48.5 & - & 31.0 \\ \hline
        BLIP \cite{li2023blip2} & 7B & 50.4 & 50.0 & 25.0 \\
        InstructBLIP \cite{instructblip}    & 7B & 50.4 & \color{blue}{57.0} & 26.8 \\ \hline
        \multirow{2}{*}{LLaVA \cite{llava}} & 7B &  49.0 &  \color{blue}{49.6} &  63.0 \\ 
                                            & 13B & 50.5 &  49.0 &  \color{blue}{73.0} \\                            
        \bottomrule
        \end{tabular}
    \end{subtable} 
    \label{table:counting}
\end{table}

\subsection{Object recognition.} 
\label{sec:objrec}

{\bf (Coarse-grained) Object Recognition.} To analyse the coarse-grained object recognition capabilities of VLMs, we use two benchmark datasets --  PaintSkills \cite{DallEval} and Pascal VOC \cite{Everingham10} (Table 1).  
%
We notice that all tested VLMs perform well on coarse-grained object recognition, resulting in high performance ($>$90$\%$) across all three spaces -- {\em visual} and {\em text} latents, as well as {\em response} space. Specifically, we notice that visual  and text latents have very similar performance, but there is a minor dip in the response space performance compared to the other two spaces. Note that for CLIP there is no text latent space and the response performance is obtained by matching class embeddings and image encoding in the latent space. 
The performance of BLIP class of models is lower in the response space, compared to others. We attribute this to no (BLIP) or weaker (InstructBLIP) instruction tuning. Evidence for this can be seen by comparing BLIP to its instruction-tuned counetrpart (InstructBLIP) which results in substantial  performance increase, but still does not reach LLaVA \cite{llava} on Pascal. 
\vspace{-2pt}

{\bf (Fine-grained) Object Recognition.} We find that even though VLMs perform well on visual latents and text latents, there is a drastic decrease in performance 
across both datasets (Table 2). This observation reinforces the findings in \cite{finer}, which show that overall VLMs are less capable in fine-grained recognition. However, we go a step further and make an important observation that the visual encoders ({\em i.e.}, CLIP visual latents) perform well ($>$90 $\%$) for fine-grained classes, with substantial drop in performance observed only in the responce space (minor drop in the text space). Contrary to previous works \cite{tong2024eyes} which find that VLMs deficiencies are due to visual encoder, here we find that even though the visual encoder performed well on given task and knowledge was processed effectively through visual and text latent spaces, the discriminative information did not effectively flow from text latents to response space which lead to a sharp reduction in performance. Note that previous findings \cite{finer} suggest that (text-only) language decoders themselves have information about fine-grained classes and are good at classification. Hence, we conjecture that the drastic dip in performance is due to ineffective training/fine-tuning of the projection layer which maps from text latents to the language decoder.

{\bf Object Recognition Discussion and Implications.}
Given our findings, we posit that the data used to fine-tune VLMs projection layer did not contain enough samples (was not representative) of fine-grained classes. While the encoding of the fine grain classes could be seen as being preserved in intermediate text space, it is clear that representations in that space do not align well with those in the language decoder. We posit that improvements in fine-grained recognition of VLMs can be achieved by better training of the respective projection layers. This can be achieved by either collecting image-caption datasets with finer grain of detail (e.g., most images of dogs tend to have caption mentioning the ``dog" rather than specific ``breed") and/or by focusing on hierarchical losses where miss prediction of of word representing a sub-category is more heavily penalized than that of the broader category name.

\subsection{Object Counting and Spatial Understanding}

In this section, we consider the object counting and spatial arrangement understanding tasks.
These tasks are more difficult than object recognition.
For example, in order to count, a model has to recognize objects and then group similar/same objects to get the total number of instances of a particular object of interest; for spatial understanding , the model has to recognize the objects for which spatial arrangement is queried and then understand the orientation of those specific objects to answer the question. Given that these tasks are more difficult, one would naturally expect performance of VLMs on them to be lower. 
Following previous work \cite{DallEval}, we consider the PaintSkills diagnostic evaluation dataset here as it allows us to control object placement and arrangement. Specifically,  we use images with 1 through 4 object instances for counting task and 4 orientations ({\em left}, {\em right}, {\em above}, {\em below}) for the spatial arrangement task. The random guess performance on both tasks is $25$\%.

{\bf Object Counting.} The results reported in Table~\ref{table:counting} (top) illustrate trends similar to those observed on fine-grain recognition, albeit less pronounced. Mainly, performance of linear probes on {\em visual} and {\em text} latent representations is quite good ($>$90$\%$), with {\em text} representations marginally trailing {\em image}, while performance in the response space is considerably poorer (a drop of at least $14$\%). 

{\bf Spatial Understanding.} Interestingly, we notice that the trend observed thus far reverses on spatial understanding (see Table~\ref{table:counting} (bottom)). The performance of
the {\em visual} latent representation is the worst, with performance increasing some in the {\em text} latent representation space (for InstructBLIP and LLaVA) and even further (for LLaVA) in the response space. Based on these results, we can make a few important deductions. First, spatial understanding capabilities of all VLMs models are clearly inferior, with image encoder being responsible for the loss of information. Second, it appears that VLMs, at least those instructionally tuned, make up for some loss of performance by leveraging language priors. 
We attribute this to the fact that likely the training data used for fine-tuning the projection layer from text latents to language decoder had samples which representative of the spatial arrangement of objects in images. Overall, we conjecture that the lack of spatial understand capabilities in VLMs is primarily due to the deficiencies of visual encoder ({\em i.e.}, CLIP) and efforts to improve the visual encoder might help improve these capabilities in VLMs.

\vspace{-5pt}
\section{Conclusion and Future Work}
\label{conclusion}
In this paper, we evaluated basic visual understanding capabilities of VLMs by going beyond final performance and conducting detailed analysis in terms of intermediate spaces that represent information as it is processed through the VLM network. In future work, we will try to evaluate more VLMs from each type/categories of VLMs considered in this paper and make our observations more comprehensive and reliable.

{\small
\bibliographystyle{ieee_fullname}
\bibliography{NLP.bib}
}

\end{document}

%% file: includes/krishnas_includes.tex
\usepackage[nolist]{acronym}
\usepackage{adjustbox}
\usepackage{amsmath}
\usepackage{amssymb}
\usepackage{amsthm}
\usepackage{array}  
\usepackage{booktabs}
\usepackage{capt-of}
\usepackage[font={small}]{caption}
\usepackage{subcaption}
\usepackage{floatrow}
\usepackage{colortbl}
\usepackage{wrapfig}
\usepackage[table,xcdraw]{xcolor}
\usepackage{graphicx}
\usepackage{multirow}
\usepackage{paralist}
\usepackage{tabularx}
\usepackage[table, xcdraw]{xcolor}
\usepackage{listings}

\definecolor{mathpurple}{rgb}{0.73, .58, .92}
\definecolor{stringsgreen}{rgb}{.5, .72, 0.42}
\definecolor{entityyellow}{rgb}{1 .83, .15}
\definecolor{relationblue}{rgb}{.42, .64, .82}

%% file: includes/krishnas_macros.tex
\makeatletter
\def\maxwidth#1{\ifdim\Gin@nat@width>#1 #1\else\Gin@nat@width\fi}
\makeatother

\definecolor{flodarkpurple}{rgb}{0.288,0.1196,0.7}

\definecolor{amber}{rgb}{1.0, 0.75, 0.0}

